\documentclass[review]{elsarticle}

\usepackage{hyperref}

\usepackage{hyperref}
\usepackage{times}
\usepackage{epsfig}
\usepackage{graphicx}
\usepackage{amsmath}
\usepackage{amssymb}
\usepackage{amsthm}
\usepackage{breqn}
\usepackage[bottom]{footmisc}

\usepackage[]{algorithm}
\usepackage[noend]{algpseudocode}

\usepackage{latexsym}
\usepackage{url}
\usepackage{multirow}

\makeatletter
\def\BState{\State\hskip-\ALG@thistlm}
\makeatother

\journal{Journal}

\begin{document}

\begin{frontmatter}

\title{Probabilistic framework for solving Visual Dialog}

\author{Badri N. Patro$^1$ , Anupriy$^2$, Vinay P. Namboodiri$^2$}
\address{$^1$ Department of Electrical Engineering, Indian Institute of Technology Kanpur, India\\ $^2$ Department of Computer Science and Engineering, Indian Institute of Technology Kanpur, India }
\ead{(badri,anupriy,vinaypn)@iitk.ac.in}




\begin{abstract}
	In this paper, we propose a probabilistic framework for solving the task of `Visual Dialog'. Solving this task requires reasoning and understanding of visual modality, language modality, and common sense knowledge to answer. Various architectures have been proposed to solve this task by variants of multi-modal deep learning techniques that combine visual and language representations. However, we believe that it is crucial to understand and analyze the sources of uncertainty for solving this task. Our approach allows for estimating uncertainty and also aids a diverse generation of answers. The proposed approach is obtained through a probabilistic representation module that provides us with representations for image, question and conversation history, a module that ensures that diverse latent representations for candidate answers are obtained given the probabilistic representations and an uncertainty representation module that chooses the appropriate answer that minimizes uncertainty. We thoroughly evaluate the model with a detailed ablation analysis, comparison with state of the art and visualization of the uncertainty that aids in the understanding of the method. Using the proposed probabilistic framework, we thus obtain an improved visual dialog system that is also more explainable.

\end{abstract}


\end{frontmatter}

\section{Introduction}\label{intro}

\begin{figure*}[!htb]
	\centering
	\includegraphics[width=1.0\textwidth]{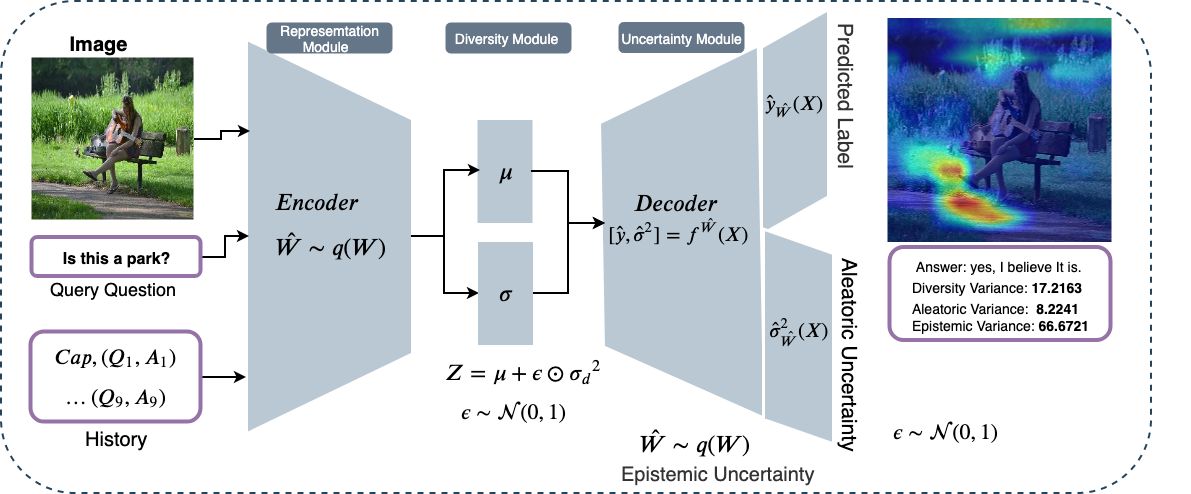}
	\vspace{-3em}
	\caption{Proposed Probabilistic Diversity and Uncertainty Network (PDUN) consists of three parts, viz. a) Probabilistic Representation Module encodes image feature with a question and history feature in an attentive manner. b) Diversity module captures the diversity, and diverse answer is generated using Variational Auto-Encoder. c) Uncertainty module predicts uncertainty of the network.}
	\label{fig:intro}
\end{figure*}

\begin{figure}[ht]
	\centering
	\includegraphics[width=0.95\textwidth]{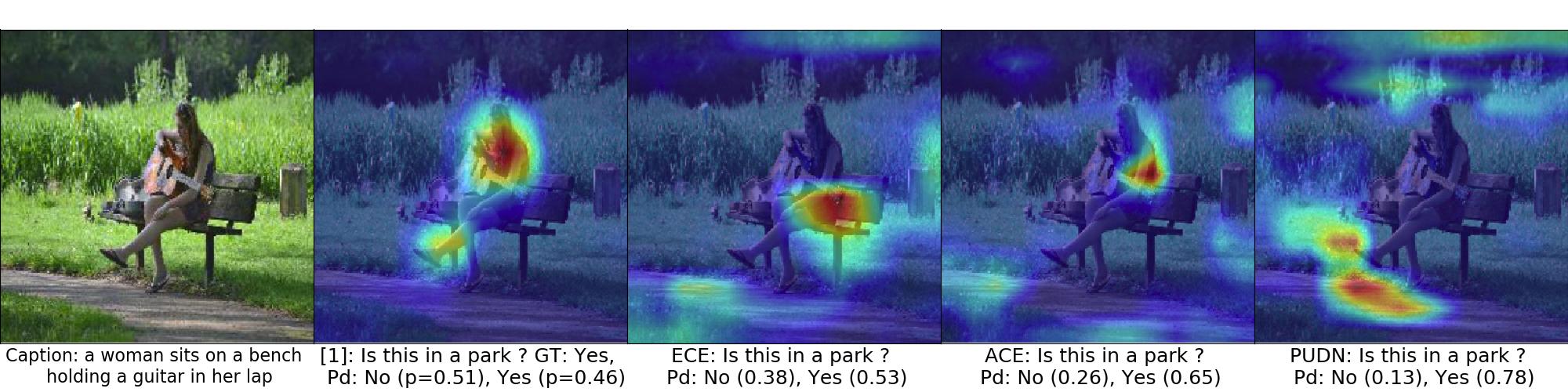}
	\vspace{-1em}
	\caption{Results were showing the certainty of the correct class increases from baseline\cite{Das_ICCV2017} to our proposed uncertainty model (PDUN). In this figure, we show the top 2 class confidence score of the question, "Is this a park?". In the baseline model focus on woman, guitar and chair and predicts "NO," which is confused with the correct prediction of the answer, whether it is a park or not. PDUN model minimizes the uncertainty and predicts the correct answer "Yes" with a high confidence score.}
	\label{fig:result_0_B}
\end{figure}

Deep learning has aided significant progress in solving various computer vision tasks such as object classification \cite{Simonyan_arXiv2014,he_CVPR2016deep} and object detection \cite{ren_NIPS2015faster,zhao_TNNLS2019object}. The solution of more semantic tasks such as visual question answering \cite{VQA,goyal_CVPR2017making} and image captioning \cite{Vinyals_CVPR2015,Karpathy_CVPR2015} has also seen progress lately. A challenging problem that extends these is that of maintaining a dialog with a user \cite{Das_ICCV2017,Das_CVPR2017} 
In this case, a system is required to maintain context concerning the history of the conversation while answering a question and this is be more challenging. A specific task in the visual context is that of the `Visual Dialog' task \cite{Das_CVPR2017}. The aim here is that given an image, we need to train an agent to maintain a dialog. The motivation for this emerges from an interest in developing associative technologies for visually impaired persons or chat-bot based dialog agents. Several methods have been proposed for solving the task, such as using various discriminative and generative encoder-decoder frameworks that aim to solve the task of generating dialog ~\cite{Das_CVPR2017, Das_ICCV2017}. In this paper we aim to extend the previous approaches by formulating a probabilistic approach towards solving this task. This approach is illustrated in figure~\ref{fig:intro}. Through our approach we can obtain a principled model that we can train end-to-end while being able to have uncertainty estimates and the ability to evaluate and explain the model. Such an ability to explain the model is crucial, especially, when we consider that the method could be used by visually impaired people. At any point in the method, the model can be probed to ensure that it is certain about the answers that it generates and more importantly any failure of the method can be addressed by explaining the precise reason for failure as shown in ~\ref{fig:result_0_B}. However, as the task is challenging it is important to have an insight into obtaining estimates regarding the uncertainty of the model. This would aid in knowing when the method is confident about its prediction. In this paper, we consider the task of understanding the uncertainty while solving the `Visual Dialog' task, as shown in ~\ref{fig:result_0_B}. This proposed method addresses the limitations of the previous approaches as the previous approaches do not have the ability to obtain uncertainty estimates or obtain diverse conversations. 
Our method consists of the following parts :
\begin{itemize}
	\item Probabilistic Representation Module: Through this module, we obtain probabilistic representations for image, question, and history of the conversation using Bayesian CNN and Bayesian RNN modules. 
	\item Diverse Latent Answer Generation Module: In this module, we use a variational autoencoder based latent representation that allows us to obtain latent representations from which we can sample answers.
	\item Uncertainty Representation Module:  In this module, we propose a Reverse Uncertainty based Attention Map (RUAM) method by using Bayesian deep learning methods that allows us to minimize data uncertainty and model uncertainty. 
\end{itemize}

To provide an overview of the technical contributions we make, the main idea is to consider incorporating a Gaussian prior for generating samples of answers. We minimize the KL divergence between the prior and the posterior distribution. The other contribution is to explicitly incorporate a loss to ensure that the correlation between different samples is minimal. We further use these losses along with a loss to minimize the uncertainty. A similar loss has been considered in another context by Patro {\it et al.} \cite{Patro_ICCV2019}. In this work we are interested in a principled framework for minimizing uncertainty by sampling and generating diverse answers. Moreover, its use has not been considered for the problem of visual dialog. We evaluate each of the contributions in our work. The technical details mentioned here are discussed further in detail in the following sections.

\vspace{-1em}

To summarize, the main contributions of this paper are as follows :
\begin{enumerate}
	\item We provide a module to obtain a probabilistic representation for image, question and conversation history that are obtained as input. 
	\item The probabilistic representations are used to {\it{generate}} diverse latent representations for candidate answers
	\item We propose a method for obtaining reverse uncertainty based on attention maps (RUAM). These allow us to select an appropriate answer that minimizes uncertainty.
	\item We provide extensive analysis and comparison of our framework with previous methods and evaluate the various ablations of the method. Our proposed framework improves recall@10 score by 3.5\%,  mean reciprocal rank (MRR) by 1\% and NDGC score by ~1\%.  Moreover, we also  provide estimation \& visualization of the uncertainty of the output.
\end{enumerate}

\section{Related Work}
\label{sec:lit_surv}

A novel problem, `Visual dialog' has been introduced recently \cite{Das_ICCV2017}. In this problem, we aim to answer a question given the context of a conversation about an image. This is one of the latest challenging problems that has been posed in the field of vision and language. 
There have been various other related problems considered in the field of vision and language. One of the earliest such problems is that of image captioning where we aim to generate a sentence describing an image \cite{Barnard_JMLR2003,Farhadi_ECCV2010,Kulkarni_CVPR2011,Socher_TACL2014,Vinyals_CVPR2015,Karpathy_CVPR2015,Fang_CVPR2015,Johnson_CVPR2016,Yan_ECCV2016}. Further, the community moved on to answer questions based on an image in the visual question answering (VQA) task \cite{Malinowski_NIPS2014,VQA,Ren_NIPS2015,Noh_CVPR2016,Xu_ECCV2016,Lu_NIPS2016,Shih_CVPR2016,Patro_CVPR2018dvqa,kim_NIPS2018bilinear,anderson_CVPR2018bottom,bai_ECCV2018deep,Ben_ICCV2017,zhang_ICLR2018learning}. Another interesting problem that has been addressed is that of visual question generation (VQG)~\cite{Mostafazadeh_ACL2016,jain_CVPR2017,patro_EMNLP2018multimodal,patro_COLING2018learning}, where the aim is that given an image to generate natural questions similar to that asked by humans. 
The solution of the visual dialog problem builds up on the previous work conducted for solving the various problems described above. 

Visual dialog task requires the agents to have meaningful dialog conversation about the visual content. This task was introduced by Das {\it et al.} \cite{Das_CVPR2017}. The authors have proposed three approaches, namely late fusion in which all the history rounds are concatenated, attention-based hierarchical LSTM which handles variable length history and memory-based method for performing results best in terms of accuracy for solving this task. Following up, \cite{lu2017best,wu_CVPR2018} have proposed generator and discriminator based architecture. Of these, Lu {\it et al.} \cite{lu2017best} consider an attention based method to combine all history rounds to get a single representation. Further works \cite{de2017guesswhat,strub2017end,Das_ICCV2017} have proposed visual dialog as an image guessing game. The latest work on the visual dialog to obtain state of the art results has been proposed by  Jain {\it et al.} \cite{jain2018two}. This work is based on discriminative question generation and answering.  In another work, Jain {\it et al.} \cite{jain_CVPR2017} have proposed a method to bring diversity in the question generation from an image using Variational Auto-encoder (VAE). Wang {\it et al.} \cite{wang_NIPS2017diverse} have proposed a similar kind of method to generate a caption from an image using VAE. In related works, \cite{dai_ICCV2017towards, li_arxiv2018generating} have captured diversity in the caption generation from an image using generative adversary network.  In contrast to these earlier works, in our framework we consider a fully probabilistic framework for solving the task of visual dialog. 

We use Bayesian CNN in our work for obtaining probabilistic image representations. Modeling distribution over CNN filters is still a difficult task. Due to the large number of parameters to be inferred, the posterior distribution becomes intractable. To approximate this posterior distribution, the variational inference is one of the existing approaches introduced by ~\cite{Hinton_ACM1993, Barber_NATO1998, Graves_NIPS2011, Blundell_ARX2015}.
Gaussian distribution is the simplest variational approximation used to fit the model parameters to the true distribution of parameters, but it is computationally expensive~\cite{Blundell_ARX2015}. This can be overcome using Bernoulli approximation. 

There has been some work done in terms of estimating uncertainty in the predictions using deep learning — the work by \cite{Gal_ICML2016} estimates the predictive variance of the deep network with the help of dropout \cite{srivastava2014dropout}. \cite{kendall2015bayesian} has proposed a method to capture model uncertainty for image segmentation task. They observed that softmax probability function approximates relative probability between the class labels, but does not provide information about the model's uncertainty. Recently, \cite{Kendall_NIPS2017} has decomposed predictive uncertainty into two major types, namely aleatoric and epistemic uncertainty, which capture uncertainty\cite{smith2018understanding,teye2018bayesian} about the predicted model and uncertainty present in the data itself. Here, our objective is to generate diverse answer, to analyze and minimize the uncertainty in answer data, and to analyze the uncertainty of the model for the challenging visual dialog task. We build up on the techniques proposed in several such works to obtain a fully probabilistic framework for solving the visual dialog problem. In the next section we consider the background in terms of Bayesian modeling required for obtaining our probabilistic framework.

\section{Background: Bayesian Approach of Model}
Consider the distribution $p(x,y)$ over the input features $x$ and labels $y$. For the visual dialog classification task, $x$ corresponding to joint encoding feature of image, history and query question and $y$ answer class label. For the given observation, $X$ and its corresponding output $Y$. In a Bayesian framework, the predictive uncertainty of the classification model ($p(y^{*}|x^{*},D))$\footnote{Standard shorthand notation for $p(y=y^{*}|x^{*},X,Y)= p(y^{*}|x^{*},D)$} is trained on a  finite set of training data $D={\{x_i,y_i\}}_{i=1}^{N}$. The predictive uncertainty will result in data(aleatoric) uncertainty and the model(epistemic) uncertainty. The model estimates two kinds of uncertainty, i.e., data uncertainty and model uncertainty. The posterior distribution describes the data uncertainty over class labels, given set of model parameters $w$, and the model uncertainty is described by the posterior distribution over model parameters $w$, given input data. The predictive uncertainty for new example point $x^*$ by integrating over all possible set of parameters w is given by 
\begin{equation}
\label{eq1}
p(y^*|x^*,X,Y)= \int{\underbrace{p(y^*|x^*,w)}_{Data}\underbrace{p(w|X,Y)}_{Model}dw}
\end{equation}
Our main objective is to find the best set of weights of our model that will generate our data X, Y. One of the approaches to make Bayesian inference is to compute the posterior distribution overweights, i.e., $p(w|X, Y)$. This distribution captures the best plausible set of model parameters given our observed data. 
\[p(w|X,Y) =  p(Y|X,W)p(W)/p(Y|X)\]
It is challenging to perform inference over the Bayesian network because the marginal probability $p(Y|X)$ of the posterior cannot be evaluated analytically. So, the posterior  distribution $p(w|X,Y)$ is  intractable. To approximate  the  intractable posterior distribution, various approximation approaches are proposed in \cite{Bishop_Springer2006,Gal_ICML2016,Blundell_ARX2015}. Variational inference is one of the approximating technique, where the posterior $p(w|X, Y)$ is approximated by a simple distribution $q_\theta(W)$, where $\theta$ is the parameterized by variational parameter $p(w|X,Y) \approx q_\theta(W)$.  We thus minimize the Kullback–Leibler(KL) divergence between approximate distribution $q_{\theta}(w)$ and the posterior $p(w|X,Y)$ w.r.t $\theta$, which is denoted by $ KL(q_{\theta}(w) || p(w|X, Y))$. 
\begin{equation*}
\begin{split}
KL(q_{\theta}(w) || p(w|X, Y)) &\propto -\int{q_{\theta}(w)} \log {p(Y|X,w)}dw + KL(q_{\theta}(w)||p(w))\\
& = -\sum_{i=1}^{N}{\int{q_{\theta}(w)} \log {p(y_i|f^{\hat{W}}(x_i))}}dw   + KL(q_{\theta}(w)||p(w))
\end{split}
\label{eq2}
\end{equation*}

Minimizing the KL divergence is equivalent to maximizing the log evidence lower bound ~\cite{Bishop_Springer2006} with respect to the variational parameters defining $q_{\theta}(w)$,
\begin{equation}
L_{VI} =  \int{q_{\theta}(w)} \log {p(Y|X,w)}dw - KL(q_{\theta}(w)||p(w))
\label{eq8}
\end{equation}
The intractable posterior problem i.e., averaging over all the weight of the BNN, is replaced by the simple distribution function. Now we need to optimize the parameter of simple distribution function instead of optimizing the original neural network's parameters. Furthermore the integral in equation \ref{eq1} (predictive posterior) is also intractable for the neural network, which is approximated via sampling using Monte Carlo dropout \cite{Gal_ICML2016} or Langevin Dynamics \cite{welling_ICML2011bayesian} or explicit ensembling \cite{lakshminarayanan_NIPS2017simple}. So we approximate the integral with Monte Carlo sampling.
\begin{equation}
\label{eq6}
\begin{split}
p(y^*=c|x^*,X,Y) &= \int{p(y^*=c|x^*,w)p(w|X,Y)dw} \\
&\approx \int{p(y^*=c|x^*,w)q_{\theta}(w)dw}\\
& \approx \frac{1}{M}\sum_{i=1}^{M}{p(y^*=c|x^*,w^{(i)})q(w^{(i)})}\\
\end{split}
\end{equation}
where $w^{(i)} \sim q(w^{(i)})$, which is modeled by the dropout distribution and M samples of $w^{(i)}$ is obtained.
, each $p(y^*|x^*,w^{(i)})$ in an ensemble ${p(y^*|x^*,w^{(i)})}^M_{i=1}$ obtained sampled from $q(w^{(i)})$.
In the following section, we have discussed Bayesian CNN and Bayesian LSTM.
\subsection{Bayesian CNN }
One way to define a Bayesian neural network \cite{Gal_ICML2016} is to place a prior distribution over neural network weights, $w=(W_{i})_{i=1}^{L}$. Given weight matrix $W_i$ and bias $b_i$ for $i$th layer, we use standard Gaussian prior distribution over the weight matrix $p_0(W_i)=\mathcal{N}(W_i;0,1)$. The variational Bayesian approximation in a Bayesian neural network can be interpreted as adding stochastic regularization in the deterministic neural network. The stochastic regularization technique is equivalent to multiplying random noise $\epsilon_{i}$ with neural network weight matrices $M_i$.
\begin{equation}
\begin{split}
&W_{i}=M_{i}\cdot diag([ \epsilon_{i,j}]_{j=1}^{K_{i}}) \\
&\epsilon_{i,j}\thicksim Bernoulli(p_{i}), i=\{1,.,L\} ,j=\{1,.,K_{i-1}\}  
\end{split}
\label{eq5}
\end{equation}
where, $\epsilon_{i,j}$ is a Bernoulli distributed random variable with probability $p_{i}$. The $diag(.)$ operator maps vectors to diagonal matrices, whose diagonal elements are the elements of the vectors. The set of variational parameters $M_i$ is now the set of matrices $\theta=\{{M_i}\}_{i=1}^{L}$. The binary variable $\epsilon_{i,j}=0$ indicates the corresponding element $j$ in the layer $i-1$ is dropped out as an input to layer $i$. 
In CNN with dropout \cite{Wu_NN2015}, the forward propagation is formulated as,
\begin{equation}
\begin{split}
&m_{k}^{i} \thicksim Bernoulli(p_{i}) \\
&\hat{a}_{k}^{i} = {a}_{k}^{i} *m_{k}^{i}\\ 
&{z}_{j}^{i+1} =   \sum_{k=1}^{n^{(l)}} Conv(W_{j}^{l+1},\hat{a}_{k}^{i})
\end{split}
\label{bay_cnn}
\end{equation}
Here ${a}_{k}^{i}$ denotes the activations of feature map $k$ $ (k=1,2,…,n^{(l)})$ at layer $l$. The mask matrix $m_{k}^{l}$ consists of independent Bernoulli variables $m_{k}^{l}(i)$. This mask is sampled and multiplied with activations in $k$th feature map at layer $l$, to produce dropout-modified activations $\hat{a}_{k}^{i}$. These modified activations are convolved with filter $W_{j}^{l+1} $ to produce convolved features ${z}_{j}^{i+1} $. The function $f$ is applied element wise to the convolved features to get the activations of convolutional layers.

\subsection{Bayesian LSTM}
The conventional LSTM is a neural network that maps LSTM state $s_t$ (at time step $t$) and input $x_t$ to a new LSTM state $s_{t+1}$, $f_l:(s_t,x_t) \rightarrow s_{t+1}$. The state of LSTM is given by $s_t=(c_t,h_t)$, where $c_t$ is a memory state and $h_t$ is the output of the hidden state. To train a input sequence of length $T$, $x_1,x_2..., x_T$, the LSTM cell is unrolled $T$ times in to a feed forward network with initial state $s_0$ and can be represented by \[ s_j=f_l(s_{j-1},x_j) \]
In Bayesian LSTM, let $p(y^*|w,x^*)$ be the likelihood of the neural network, then the posterior of the network is approximated to $q(w)$ by minimizing the variational free energy $L(w)$\cite{Gal_NIPS2016,Fortunato_Arxiv2017}.
Minimizing the variational free energy is equivalent to maximizing the likelihood  $\log p(y|x,w)$ subject to KL divergence, which regularizes the parameters of the network.
\begin{equation*}
\label{eq3}
L(w)   =-E_{q(w)}\big[\log {p(y^*_{1:T}|x^*_{1:T},w)}\big]  + KL(q(w)||p(w))
\end{equation*}

Here $\log {p(y_{1:T}|x_{1:T},w)}$ is the likelihood function of the sequence and the expectation in the previous equation is approximated by the Monte Carlo sampling. The predictive posterior for LSTM is calculated just as in equation~\ref{eq6} by, 
\begin{equation}
\begin{split}
\scriptsize
p(y^*_{1:T}|x^*_{1:T},X,Y) &= \int{p(y^*_{1:T}|x^*_{1:T},w)p(w|X,Y)dw} \\
&\approx \frac{1}{M}\sum_{m=1}^{M}{p(y^*_{1:T}|x^*_{1:T},\Hat{w}_{m})dw}
\end{split}
\end{equation}
with $\Hat{w}_{m}\sim q_{\theta}(w)$, where $q_{\theta}(w)$ is called the dropout distribution for LSTM.

\section{Methods}
Visual dialog task is introduced by \cite{Das_CVPR2017}. The visual dialog task is defined as, given image I, a caption C, a dialog history till $t-1$ rounds, $H= \{C,(q_1,a_1),....(q_{t-1},a_{t-1}) $ and the following question $q_t$ at round $t$. The objective of the visual dialog agent is to predict a natural language answer to the question $q_t$. The visual dialog problem can be solved into two possible ways; one is by using a generative model and the other by using a discriminative model. In a generative model, given the embeddings of image, history, and question($q_t$), a generative model is trained to maximize the likelihood function to predict ground truth answer sequence. The discriminative model receives embedding of an image, history, and question($q_t$) along with 100 candidate answers $A_t=\{a_t^1,...a_t^{100}\}$ and effectively learns to rank the list of candidate answers.

One aspect of previous approaches tends to be a lack of diverse generations of answers; for instance, the tendency to correlate the animal `zebra' with black and white stripes. In contrast, during conversations, a conversation is interesting if an unexpected or novel observation is raised. In our method, we hope to produce such insights. To do that we need an ability to characterize the space of all possible answers. We do that by using a Gaussian prior for the generation of answers. This allows us to generate samples of plausible answers. We then further use a diversity loss that would penalize correlations between the multiple samples. The final task then lies to choose an appropriate retort or response. To do so, we rely on minimizing uncertainty while generating the answer. We now consider the proposed approach in detail.


\begin{figure*}[ht]
	\centering
	\includegraphics[width=1.1\textwidth]{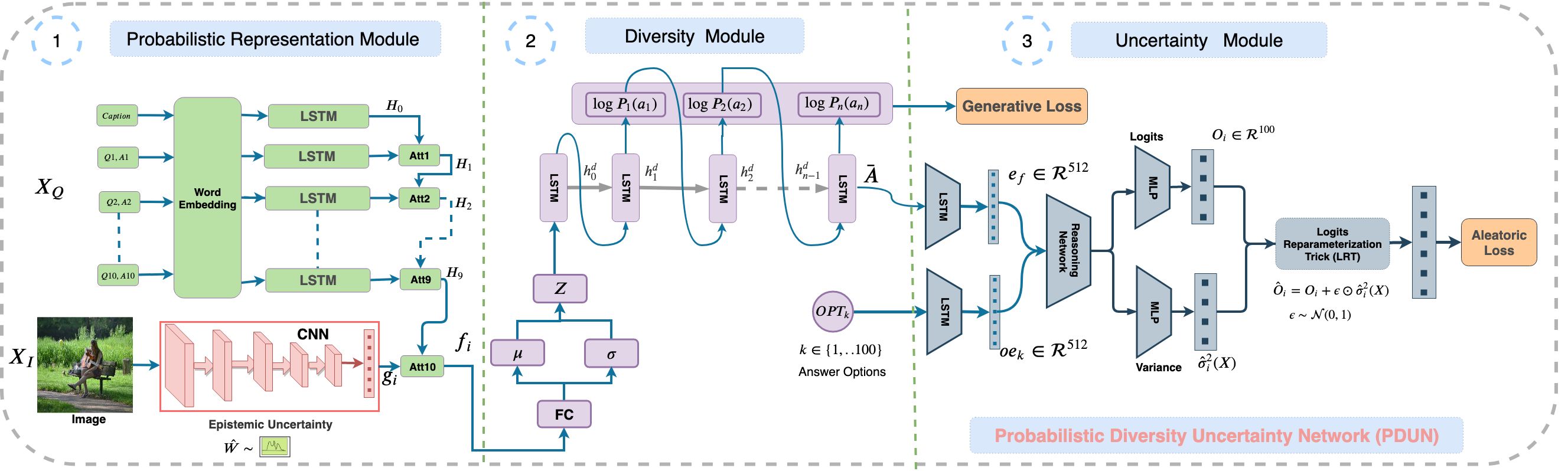}
	\vspace{-2.4em}
	\caption{Probabilistic Diversity Uncertainty Network(PDUN), Bayesian CNN/LSTM is used to obtain the embeddings $g_i,f_i,h_i$ which is then fused using the Fusion Module to get $e_f$. Then correlation is found between fused embedding with answer option embedding. Finally, variance and logits output are obtained using MLP, which is then used in Logits Reparameterization Trick(LRT) to get final softmax output.}
	\label{fig:main_figure}
\end{figure*}


Our method consists of three parts viz. 1) Probabilistic Representation Module, 2) Latent feature-based Diverse Answer Generation Module, and 3) Uncertainty Module  as illustrated in figure~\ref{fig:main_figure} :

\subsection{Probabilistic Representation Module}
We adopt a probabilistic representation module that has been previously considered in Patro {\it et al.} \cite{Patro_ICCV2019} for the VQA task. However, using this representation for visual dialog requires us to also consider the history of previous dialogs as a new input. By using a probabilistic representation, we are able to investigate the uncertainty in any part of the full proposed model. To obtain this, we use the methods of Bayesian CNN and Bayesian LSTM that has been discussed in the background section. Given an input image $x_i$, we obtain an image embedding $g_i$ by using a Bayesian CNN that we parameterized through a function $G(x_i, W_i)$, where $W_i$ are the weights of the Bayesian CNN. We extract $g_i \in \mathcal{R}^{w \times h \times c}$ dimensional CONV-5 feature from Bayesian CNN network as shown in figure~\ref{fig:main_figure}. We obtain $g_q,g_h$ encoding feature for given question and history, after passing through an LSTM (Bayesian LSTM Network), which is parameterized using the function $G_q(X_{WE}, \theta_l)$, where $\theta_l$ are the weights of the LSTM as shown in figure \ref{fig:main_figure}. Similarly, we obtain answer embedding $G_a$ parameterised by $G(X_a,\theta_a)$. After this, the question and answer embedding are combined to obtain a history embedding. To model the Bayesian CNN~\cite{Gal_ARX2015}, we use pre-trained CNN layers and put dropout layer with dropout rate $p$, before each CNN layer. Similarly for Bayesian: LSTM~\cite{Fortunato_Arxiv2017}, we add dropout on each input and a hidden layer of the LSTM cell. These are input to an attention network that combines question-answer pair with previous history embedding using a weighted softmax function and produces a weighted output attention vector $g_f$. There are various ways of modeling the attention network. In this paper, we have evaluated the network proposed in SAN~\cite{Yang_CVPR2016}.  In the last round, we combine image embedding with the last history embedding to get a dialog context vector. At each round, we attend over the question representation with the previous history (combined question-answer representation). In the first round, the previous history is an encoded caption feature. In the final round, we attend to image representation with the appropriate history representation to obtain an attentive encoder feature, $g_{f}$. The attention mechanism is illustrated as follows:
\begin{equation}
\begin{split}
& g_{a}=\tanh(W_{c}g_i+ W_{q}(g_q || g_h) + b_{c})\\
& \alpha = \text{Softmax}(W_{a}g_{a}+b_a) \\
\end{split}
\end{equation}
where $||$ means concatenation, $W_{a},W_{c},W_{q}, b_{c},b_{a}$ are the weights and bias of different layers.

\subsection{Latent feature based Diverse Answer Generation Module}
This module mainly focuses on representing a latent representative vector from attentive encoder module and generate a diverse answer using answer generator. We use the VAE ~\cite{kingma_STAT2014} based generative framework to generate diverse answer from the attentive encoder. We obtain mean, $\mu =W_{\mu}g_f $ and log variance, $\log \sigma^{2}=W_{\sigma}g_f$, where $\mu$ and $\sigma$ are the parameters of a multivariate Gaussian distribution. We train this network to learn a variational distribution which is close to a prior defined by the normal distribution with zero mean and unit variance i.e., $\mathcal{N}(0,1)$. Then we obtain a latent vector representation $\mathbf{z}$ by using the reparameterization trick $ \mathbf{z} =\mu + \epsilon \odot \sigma $. The major concern for answer generation is the spread of the variance in the latent representation. Our main objective is to increase the spread in the Gaussian as much as possible for generating diverse answers.  If the spread of the Gaussian is too low $(\sim 0)$, then we have sampled similar answers, and if the variance is too high ($\sim \infty$), then we have sampled from a uniform distribution. Hence, we want to put some constraints on variance such that our sampled latent representations are as diverse as possible. A diversity loss which minimizes the correlation between the latent representations is introduced to ensure this. Let us define $z_{1}$ and $z_{2}$ as the two latent vectors randomly sampled from the $\mathcal{N}(\mu,\sigma)$, $z_{1} =\mu + \epsilon_{1} \odot \sigma, z_{2} =\mu + \epsilon_{2} \odot \sigma $, where $\epsilon_{1}$ and $\epsilon_{2}$ are sampled from $\mathcal{N}(0,1)$ and $\mu$ and $\sigma$ are Gaussian parameters. The  diversity loss is given by 
\begin{equation}
\label{costfun_div}
L_{diverse}=\dfrac{\langle\, (z_{1}- \alpha), (z_{2}- \alpha)\rangle}{max({\vert \vert z_{1}- \alpha \vert \vert }_{2} \cdot {\vert \vert z_{1}- \alpha \vert \vert }_{2} , \gamma)}
\end{equation}
Where $\gamma =10^{-8}$ is used to avoid division by zero, and $\alpha$ is the average of all the $z$ samples. Similarly, we obtain an average loss for $k$ sample points (in our experiments, we choose $k$ to be 100) randomly sampled from the latent distribution. This loss ensures that these latent vectors are as far as possible.

Finally, the diverse latent feature is input to an LSTM based answer decoder module to generate diverse answers.
The softmax probability for the predicted answer token at different time steps is given by the following equations:
\begin{equation*}
\begin{split}
& h_0=Z_i=\mathcal{N}(\mu,\sigma) \\
& x_t=W_e*a_t,  \forall t\in \{0,1,2,...T-1\} \\
& {h_{t+1}}=\mbox{LSTM}(x_t,h_{t}), \forall t\in \{0,1,2,...T-1\}\\
& \hat{y}_{t+1} =  softmax(W_o * h_{t+1})\\
&\it{L_{CE}}= -\frac{1}{C}\sum_{j=1}^{C} y_{j} \log \texttt{p}(\hat{y}_{j}|f_o)
\end{split}
\end{equation*}
where $\hat{y}_{t+1}$ is the predicted answer class and $f_o$ is the context. Now, we classify the generated diverse answer among  100 classes in order to rank them with 100 ground truth answer options. We use a reasoning network to perform reasoning by predicting an answer and comparing it with the ground truth answer to obtain the final score. A similar approach has been used by Das {\it et al.} \cite{Das_ICCV2017}.

\subsection{Uncertainty Representation Module}
\label{Uncertainty}
Through the previous module, we obtain the ability to generate diverse answers. The task then is to choose an appropriate answer that is correct. To do that, we use an uncertainty representation module that characterizes the uncertainty among the diverse set of candidate answers (i.e., the classes present in the answers). We want to be certain about the response that is chosen. That is, we would like to minimize the uncertainty. We do that by using an explicit loss for reducing uncertainty. 

\begin{figure*}[ht]
	\centering
	\includegraphics[width=1\textwidth]{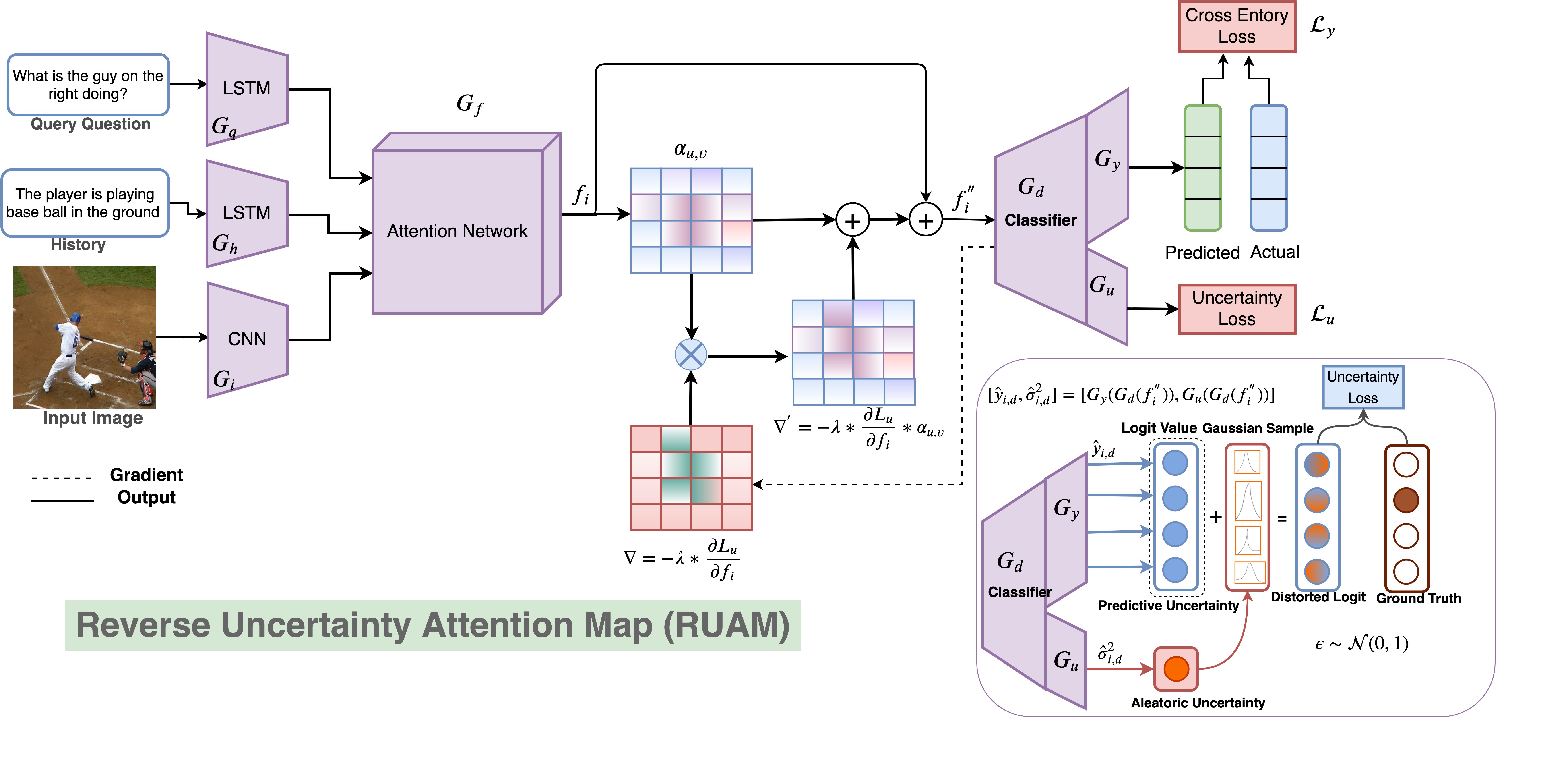}
	\vspace{-3em}
	\caption{Reverse Uncertainty based Attention Map (RUAM): We obtain attention embedding $f_i$ from the attention network $G_f$ using image, question and history embeddings $g_i,g_q,g_h$. Then we classify into answer class and obtain the uncertainty present in the data. Then we obtain reverse uncertainty map with will combine with attention map to get better confidence on the attention map as shown in the figure.}
	\label{fig:rum}
\end{figure*}

In this work, we also incorporate the attention regions, which specifies the spatially distributed weights given to a specific region embedding while generating the answer. To obtain the best embedding, we consider the ground-truth answer and through attention, consider the corresponding spatial location. This region is multiplied with the uncertainty for generating the spatial attention weighted uncertainty corresponding to the ground-truth answer. We want to increase the weight for the spatial attention corresponding to generating the ground-truth answer and minimize the uncertainty for the same. At the same time, we would like to increase the uncertainty for all other answers and minimize the weight given in terms of attention to all other regions. We achieve this through a reverse uncertainty based attention map (RUAM) that is shown in figure~\ref{fig:rum}. 


\textbf{Reverse Uncertainty based Attention Map (RUAM): }
Patro {\it et al.}\cite{Patro_ICCV2019} have proposed a model to estimate aleatoric and predictive uncertainty for Visual Question Answering task, where the gradient of uncertainty loss and gradient of classification is multiplied to improve attention feature. Kurmi {\it et al.} \cite{Kurmi_CVPR2019attending} have also proposed a similar kind of network in the domain adaption task, where they train the discriminator network to reduce uncertainty in source and target domain. We follow a similar type of network in a visual dialog task to reduce uncertainty in the attention mask with the help of a predicted answer in the dialog turns. We stress more on those attention regions whose uncertainty is less and vice-versa. The aleatoric uncertainty occurs due to corruption in the feature or noise in the attention regions. These regions are the main source of predicting the wrong answer in the visual dialog.

We adopt a Bayesian framework to predict answer classification uncertainty efficiently. We make our answer classifier as Bayesian and perform probabilistic inference over the classifier to obtain the final answer score. We adopt a Bayesian classifier as considered in several works ~\cite{Gal_NIPS2016,Gal_ICML2016, Fortunato_Arxiv2017,Kurmi_CVPR2019attending,Patro_ICCV2019}. The Bayesian classifier is obtained by applying dropout after every fully connected (FC) layer in the classifier and the Bayesian classifier predicts answer class logits $y_{i,d}$ and aleatoric uncertainties. These are obtained as follows:
\begin{equation}
\label{e5}
y_{i,d}= G_y(G_d(f_i)), \quad \sigma_{i,d}= G_v(G_d(f_i)) 
\end{equation}
where $f_i$ is the attention feature for input image sample $x_i$, question sample $x_q$ and history sample $x_h$, which is obtained by the attention feature extractor $G_f : f_i = G_f (G_i(x_i),G_q(x_q), G_h(x_h)$, where $G_y$ and $G_v$ are the logits and aleatoric variance predictor of the classifier $G_d$ respectively. 
The whole model is trained with the help of uncertainty loss (More details are present in ~\ref{costfun_ale}) and cross-entropy loss. The uncertainty loss helps the classifier to make the classifier features more robust for prediction. Finally, we measure the uncertainty for our answer prediction and found it to be lower. 

We learn and estimate observational noise parameter $\sigma_{i,d}$  to capture the uncertainty present in the input data (Image, History and Question). This can be achieved by corrupting the logit value ($y_{i,d}$) with the Gaussian noise with variance $\sigma_{i,d}$ (diagonal matrix with one element for each logits value) before the softmax layer. We used a Logit Reparameterization Trick (LRT) \cite{Gal2016Uncertainty}, which combines two outputs $y_{i,d},\sigma_{i,d}$ and then we obtain a loss with respect to ground truth. That is, after combining we get $\mathcal{N}(y_{i,d},(\sigma_{i,d})^2)$ which is expressed as:
\begin{equation}
\label{e4}
\hat{y}_{i,t,d}=y_{i,d}+\epsilon_{t,d}\odot  \sigma_{i,d},\quad where \quad \epsilon_{t,d} \sim \mathcal{N}(0,1)
\end{equation}
\begin{equation}
\label{e6}
\mathcal{L}_u= \sum_{i}\log \frac{1}{T}\sum_{t}{\exp{(\hat{y}_{i,t,M} - \log \sum_{M^`}\exp{\hat{y}_{i,t,{M^`}}})}}
\end{equation}
where $M^{'}$ is a discrete word token present in each sample sentence.  $y_{i,t}$. M is a discrete word token present in the ground truth sentence, $\mathcal{L}_v$ is minimized for the true work token $M$, and t $\in$ T is the number  of  Monte  Carlo  simulations. $\sigma_{i,d}$ is the standard deviation, ( $\sigma_{i,d}=\sqrt{v_{i,d}}$). 

Now, we obtain uncertainty for attention map $\alpha_{att} \in  \mathcal{R}^{u \times v}$ of width u and height v using following steps such as, we first compute gradient of the predictive uncertainty $\sigma_g^2$ of our generator with respect to the features $f_i$. This gradient of the uncertainty loss  $\mathcal{L}^u$ with respect to the attention feature $f_i$ is given by $\frac{\partial{\mathcal{L}_v}}{\partial{f_i}}$. Now we pass the uncertainty gradient through a gradient reversal layer to reverse the gradient to get certainty mask for the attention map. This is given by \[\nabla_{u} = -\gamma * \frac{\partial{\mathcal{L}_u}}{\partial{f_i}}\]
We perform an element-wise multiplication of the forward attention feature map and reverse uncertainty gradients to get an enhanced attention feature map i.e.
\begin{equation} \alpha^{'}_{u,v}= -\gamma * \frac{\partial{\mathcal{L}_u}}{\partial{f_i}}* \alpha_{u,v} \end{equation}
The positive sign of the gradient $\gamma$ indicates that the aleatoric certainty is activated on these regions and vice-versa. We apply a ReLU activation function on the product of gradients of the attention map and the gradients of aleatoric certainty as we are only interested in attention regions that have a positive influence for a corresponding  answer class, i.e. attention pixels whose intensity should be increased in order to increase $y^c$, where negative values are multiplied by $\gamma$ (large negative number). Negative attention pixels are likely to belong to other categories in the image.  
\begin{equation}\alpha_{u,v}^{''} =ReLU(\alpha_{u,v}^{'})+\gamma ReLU(-\alpha_{u,v}^{'})\end{equation}
Images with higher aleatoric uncertainty correspond to lower certainty. Therefore the certain regions of these images should have lower attention values.  We use residual connection to obtain the final attention feature by combining original attention feature with the reverse uncertainty map $\alpha_{u,v}^{''}$. This is given by:
\begin{equation}
\begin{split}
\alpha_{new}&= \alpha_{u,v} + {\alpha^{''}_{u,v} *\alpha_{u,v}}  \\
f_i^{'} &= \sum_{u,v}{g_{i} *\alpha_{new}} 
\end{split}
\end{equation}
Where, $g_i \in G_i(x_i)$. The final attention feature $(f_i^{''})$ can be obtained by combining attention feature $(f_i)$ with RUAM based new attention feature $(f_i^{'})$.
\begin{equation}
f_i^{''} = f_i+  f_i^{'}
\end{equation}
We show here, that using reverse uncertainty based attention Map (RUAM) results in an improved attention network and the attention confidence also increases. The entropy and predicted variance of the sampled logit's probability can be calculated as:
\begin{equation}
\label{eq10}
H(\hat{y}_{i,t})= -\sum_{m=1}^{M}{p(\hat{y}_{i,t}=M)*\log{p(\hat{y}_{i,t}}=M)}
\end{equation}
The predictive uncertainty is the combination of entropy and variance of $T$ sample outputs (of randomly masked model weights).
\begin{equation}
\label{eq11}
\sigma^2_p =\frac{1}{T}\sum_{t=1}^{T}{H(\hat{y}_{i,t})} + \frac{1}{T}\sum_{t=1}^{T}{v_{i,t}^{2}}
\end{equation}
Where $H(\hat{y}_{i,t})$ is the entropy of the probability $p(\hat{y}_{i,t}^c)$, which depends on the spread of the probabilities and the variance captures both the spread and the magnitude of outcome values $\hat{y}_{i,t}$. Algorithm-\ref{alg:GCA} explains details about reverse uncertainty map for attention mask. 
\subsection{Cost Function} \label{costfun_ale}

Finally, we trained our complete PDUN model with the help of answer generation loss and uncertainty loss. The answer generation loss $L_{gen}$ is the combination of cross entropy loss $L_{CE}$, to generate each and every token in the answer sequence, KL divergence loss $L_{KL}$, to bring the approximate posterior closer to $\mathcal{N}(0,1)$, and the diversity loss $L_{div}$ ~\ref{costfun_div}, to ensure diverse answer generation.  The cost function used for obtaining the parameters $\theta_f$ of the attention network, $\theta_c$ of the classification network, $\theta_y$ of the prediction network and $\theta_u$ for uncertainty network is as follows:
\[C(\theta_f,\theta_c,\theta_y,\theta_u)=\frac{1}{n}\sum_{j=1}^{n}{ L^j_y(\theta_{f},\theta_c,\theta_y)}+ L^j_{KL}(\theta_{f},\theta_c) + L^j_{div}(\theta_{f},\theta_c) + \eta L^j_u(\theta_{f},\theta_c,\theta_u) \]
where n is the number of examples, and $\eta$ is a hyper-parameter that is fine-tuned using validation set and $L_c$ is standard cross entropy loss.
We train the model with this cost function till it converges so that the parameters $(\hat{\theta}_f,\hat{\theta}_c,\hat{\theta}_y,\hat{\theta}_u)$ deliver a saddle point function
\begin{equation}
\begin{split}
& (\hat{\theta}_{f},\hat{\theta}_{c},\hat{\theta}_{y},\hat{\theta}_u)= \arg\max_{\theta_f,\theta_c,\theta_y,\theta_u}(C(\theta_f,\theta_c,\theta_y,\theta_u))\\
\end{split}
\end{equation}




\begin{algorithm}[!htb]
	\caption{Reverse Uncertainty based Attention Map (RUAM) }\label{alg:GCA}
	\begin{algorithmic}[1]
		\Procedure{RUAM}{$I,Q,H$}
		\State {\bfseries Input:}  Image $X_I$, Question $X_Q$, History $X_H$
		\State {\bfseries Output:}  Answer $y$      
		\While{loop}
		\State  \text{Attention features  $ G_f(G_i(X_I),G_q(X_Q),G_H(X_H))\gets f_i$}
		\State  \text{Answer Logit $ G_y(G_d(f_i))\gets \hat{y}$}
		\State  \text{Data Uncertainty $ G_u(G_d(f_i))\gets \sigma^2_A$}
		
		\State  $\sigma^2_W=\sigma^2_A + H(\hat{y}_{i,t}) , \text{(Ref: eq-~\ref{eq5})}$
		\State  \textit{Ans cross entropy $\mathcal{L}_y \gets$ loss$(\hat{y},y)$}
		\State Variance Equalizer~\cite{dorman_github2014}  $\mathcal{L}_{VE}: =\sum{ReLU(\exp^{\sigma^2_{w}} - \exp^{I}}) $, 
		\While{$t=1: \# MC-Samples$}
		\State \textit{Sample ${\epsilon}_t^{w} \sim \mathcal{N}(0,\sigma^2_W)$ }
		\State \textit{Distorted Logits:$ \hat{y}_{i,t} ={\epsilon}^{w}_t + \hat{y}_i $}
		\State \textit{Gaussian Cross Entropy ~\cite{dorman_github2014} $\it{L_{GCE}}= -{\sum_{}^{} y \log \texttt{p}(\hat{y_d}|F(.))}$} 
		\State \textit{Distorted Loss :$\it{\mathcal{L}_{UDL}}= \exp(\it{\mathcal{L}_{y}}-\it{\mathcal{L}_{GCE}})^2$}
		\State \textit{Aleatoric uncertainty loss $\it{\mathcal{L}_{u}}= \it{\mathcal{L}_{GCE}}+\it{\mathcal{L}_{VE}+\mathcal{L}_{UDL}}$}
		\EndWhile
		\State \textit{Compute Reverse Gradients w.r.t $f_i$, $\nabla_u= -\lambda* \frac{\partial \mathcal{L}_{U}}{ \partial  f_i}$}
		\State \textit{Certainty Activation  for attention $\alpha_{u,v}^{'}= \nabla_u * \alpha_{u,v}$}
		\State \textit{Certainty Activation  for attention $\alpha_{u,v}^{''}= ReLU(\alpha_{u,v}^{'}) + \gamma* ReLU(-\alpha_{u,v}^{'})$}
		\State \textit{New Attention gradient $\alpha_{new}=\alpha_{u,v}+\alpha_{u,v}^{''} * \alpha_{u,v} $}
		\State \textit{New attended feature: $ f^{'}_{i}= \sum_{u,v}{f_{i} * \alpha_{new}}$}
		\State \textit{Final attended feature: $ f^{''}_{i}= f_{i} + f^{'}_{i}$}
		\State  \textit{update $\theta_f \gets \theta_f - \eta \nabla_y^{'}$} 
		\EndWhile
		\EndProcedure
	\end{algorithmic}
\end{algorithm}

\begin{figure}[ht]
	\centering
	\begin{tabular}[b]{ c}
		\includegraphics[width=0.85\textwidth]{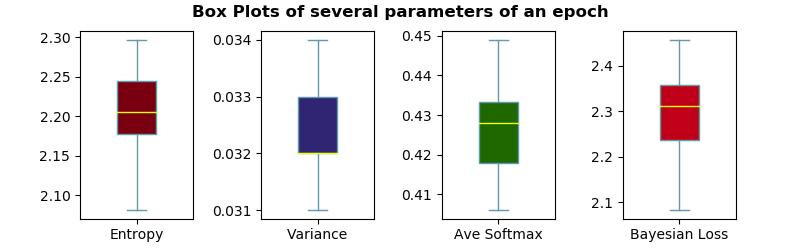}\\
		\includegraphics[width=0.85\textwidth]{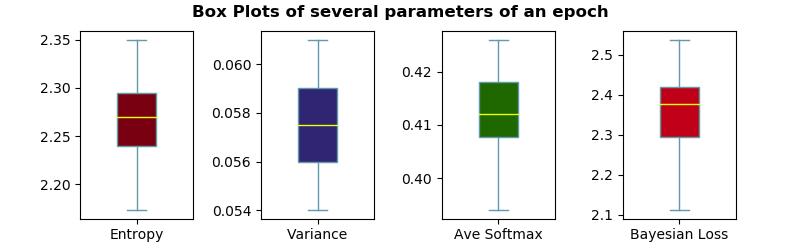}
	\end{tabular}
	\vspace{-1em}
	\caption{This shows the measurement of Entropy, Variation, Softmax scores and Bayesian loss for Bayesian model with dropout value 0.5 and 0.1 in first and second plot respectively for capturing Epistemic Uncertainty}
	\label{fig:result_1_A}
\end{figure}

\section{Experiments}
\begin{figure}[!htb]
	\small
	\centering
	\begin{tabular}[b]{ c}
		\includegraphics[width=0.8\textwidth]{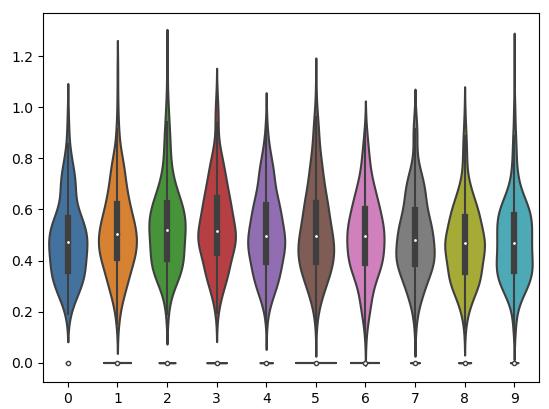}\\
		\includegraphics[width=0.8\textwidth]{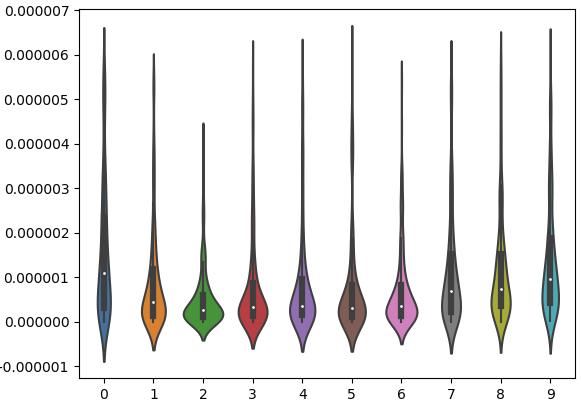}
		\vspace{-0.72em}
	\end{tabular}
	\caption{Upper Graph: This shows how aleatoric uncertainty loss varies over different turns in visual dialog. There is an eventual decreasing trend. Lower Graph: This shows how variance equalizer loss varies over different turns in visual dialog. There is an eventual increasing trend.}
	\label{fig:result_ale}
\end{figure}

{\label{experiments}}
We evaluate the proposed method in the following steps: First, we evaluate our proposed uncertainty model against other variants described in section~\ref{variants}. Second, we have shown analysis results for epistemic uncertainty in figure-\ref{fig:result_1_A} and aleatoric uncertainty in figure-\ref{fig:result_ale} and in table \ref{tab_Uncertainty}. Third, we further analyze effect of noise in aleatoric and epistemic uncertainty in table \ref{tab_gamma_noise_ALE_EPI}. Fourth, we compare diversity score for different variants of our model in table \ref{tab_diversity}. Fifth, we compare our network with state-of-the-art methods such as `visdial' \cite{Das_CVPR2017} in table ~\ref{score_tab_1}. Then, we have shown the Grad-CAM \cite{selvaraju2017grad} visualization of activation due to aleatoric uncertainty and baseline model (late fusion). We further compare our network with state-of-the-art methods such as visdial~\cite{Das_CVPR2017} model. Finally, we have provided some qualitative results of our visual dialog model. 
The quantitative evaluation is conducted using standard retrieval metrics, namely (1) mean rank, (2) recall @k, (3) mean reciprocal rank (MRR) of the human response in the returned sorted list.

\begin{table}[!htb]
	\centering
	\begin{tabular}{|p{1.4cm}|c|c|c|c|c|}
		\hline
		\textbf{Loss}&  \textbf{R1} & \textbf{R5} & \textbf{R10}& \textbf{MRR} & \textbf{Mean}\\
		\hline
		Baseline & 40.9& 72.4&82.8 &0.550 & 05.95\\
		VE      &30.9&51.4 &62.5&0.371& 12.41\\
		CE      &41.1& 72.6&82.9 &0.556 &05.98\\
		GCE     &43.8&77.2 &88.0&0.587& 04.79\\\hline
		CE+VE   &42.3&74.6 &84.7&0.551& 05.50\\ 
		VE+GCE  &44.3&79.5 &89.6&0.599& 04.41\\
		CE+GCE  &45.4&80.6 &91.2&0.610& 03.95\\ \hline
		ACE     &47.0& 82.4 &92.3 &0.622 & 03.81\\\hline
	\end{tabular}
	\caption{\label{tab_var_loss} Aleatoric loss variant in VisDial-v1.0 in test-std dataset.}
\end{table}

\subsection{Dataset}
We evaluate our proposed approach by conducting experiments on  Visual Dialog dataset~\cite{Das_CVPR2017}, which contains human annotated questions based on images of MS-COCO dataset.  This dataset was developed by pairing two subjects on Amazon Mechanical Turk to chat about an image. One person was assigned the job of a `questioner' and the other person act as an `answerer'. The questioner sees only the text description of an image which is present in caption from MS-COCO dataset. The image remains hidden to the questioner. Their task is to ask questions about this hidden image to ``imagine the scene better''. The answerer sees the image and caption and answers the questions asked by the questioner. The two of them can continue the conversation by asking and answering questions for 10 rounds at max. We have performed experiments on ``VisDial 1.0'' version of the dataset.
``VisDial v1.0'' contains 123k dialogs on COCO-train and 2k on ``VisualDialog\_val2018'' images for val and 8k on ``VisualDialog\_test2018'' for test-standard set. The caption is considered to be the first round in the dialog history. 

\begin{table}
	\centering
	\begin{tabular}{|p{4.2cm}|c|c|}
		\hline
		\textbf{Type of Uncertainty}&  \textbf{Mean} & \textbf{Std} \\
		\hline
		Aleatoric (with CE)       &0.0051  & 8.677\\
		Aleatoric (with VE)       &0.0044  & 7.353\\
		Aleatoric (with GCE)      &0.0039  & 3.431\\
		Aleatoric (with ACE)      &0.0032  & 2.119\\\hline
		Epistemic (50\% training) &0.6680  & 66.9321\\
		Epistemic (75\% training) &0.6310  & 42.8923\\
		Epistemic (100\% training)&0.5520  & 36.8110\\\hline
	\end{tabular}
	\caption{\label{tab_Uncertainty} Aleatoric \& Epistemic uncertainty measurement score.}
\end{table}

\subsection{Ablation Analysis on  Model Parameter for Uncertainty }\label{variants}
Aleatoric Cross Entropy consists of distorted (Gaussian Cross Entropy (GCE)), undistorted (Cross Entropy (CE)) loss, and Variance Equalizer (VE) loss.. The first block of the table~\ref{tab_var_loss} analyses individual loss function and its comparison is provided in that table. We use these models as our baseline and compare other variations of our model with the best single loss function. The GCE loss performs best among all the 3 losses. This is reasonable as GCE can guide the loss function to minimize the variance in the data. The second block of table~\ref{tab_var_loss} depicts the models which uses combination of the loss function as variations of our method such as GCE, VE or CE. We see an improvement of around 4\% in R@1, 8\% in R@5 score, 9\% in R@10 score and 5\% in MRR score from the baseline score. The combination of GCE loss and CE performs best among all the 3 cases. The third block takes into consideration all the loss functions ACE (GCE+CE+VE) and we see an improvement of around 6\% in R@1, 10\% in R@5 score, 10\% in R@10 score and 7\% in MRR score from the baseline score.  The behaviour of dialog turn for a particular example is shown in ~\ref{fig:result_ale}.  The first part of the figure~\ref{fig:result_ale} shows, how aleatoric uncertainty loss varies over different turns in visual dialog. As dialog progress the width of the dialog turn decreases. There is an eventual decreasing trend. The second part of the figure shows how variance equalizer loss varies over different turns in visual dialog. There is an eventual increasing trend. We can observe the third and fourth turn is more uncertain. This indicates that to have a successful dialog, it basically depend on the central part of the dialog not only start and end tuns of the dialog. 


\subsubsection{Analysis of Epistemic Uncertainty }\label{Epistemic_uncertainit}
One of the main purposes of the Bayesian deep learning is that it improves both the predictions and the uncertainty estimates of the model. We have measured uncertainty score in terms of  mean and variance for all the dialog prediction in ``val-v1.0" dataset. We have also measured uncertainty for a single dialog in the dataset. Here, we split our training data into three parts. In first part the model is trained with 50\% of the training data. Then, second part is trained by 75\% of training data and third part is trained by full dataset as shown in second block of the table \ref{tab_Uncertainty}. It is observed the epistemic uncertainty decreases as training data increases. 

\begin{table}[ht]
	\centering
	\begin{tabular}{|p{3.5cm}|c|c|}
		\hline
		\textbf{Type of Uncertainty}&  \textbf{Mean} & \textbf{Std} \\
		\hline
		Aleatoric (original)      &0.0067  & 8.956\\
		Aleatoric ($\gamma=0.8$)      &0.0123  & 11.717\\
		Aleatoric ($\gamma=1.2$)      &0.0034  & 6.353\\\hline
		Epistemic (original)      &0.671  & 70.213\\
		Epistemic ($\gamma=0.5$)      &0.701  & 71.893\\
		Epistemic ($\gamma=0.8$)      &0.646  & 69.117\\\hline
	\end{tabular}
	\caption{\label{tab_gamma_noise_ALE_EPI} Ablation study on Noise parameters.}
\end{table}

\begin{figure}[ht]
	\centering
	\includegraphics[width=1.0\columnwidth]{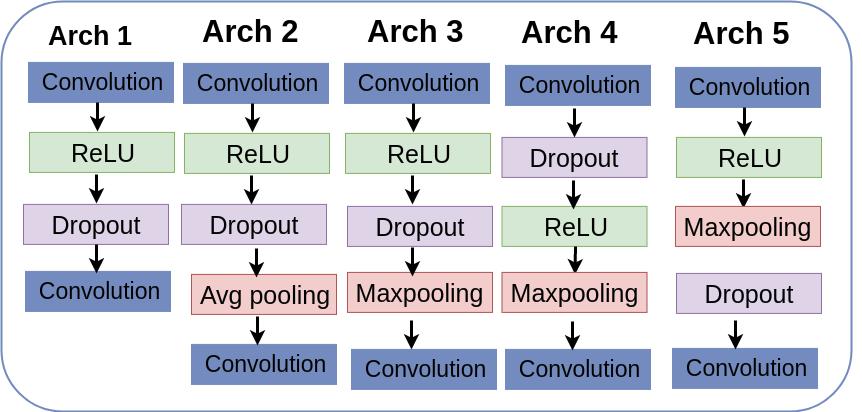} 
	\vspace{-2.2em}
	\caption{Bayesian CNN experiment based on dropout, Max-pooling and average pooling. We found out that Architecture 5 works best and we used it throughout our experiments. }
	\label{fig:univerise_1}
\end{figure}
\subsubsection{Analysis of Aleatoric Uncertainty }\label{uncertainit_variants}
Here, we have captured data uncertainty by checking contribution of each terms in aleatoric uncertainty as shown in first block of the table~\ref{tab_Uncertainty}. From the measurements, it can be easily seen that comparing aleatoric uncertainty of an image with epistemic uncertainty of another image doesn't make sense because of significant difference in their values. But both the uncertainties can be separately compared for different images to see which answer is more uncertain. 
\subsubsection{Analysis of Noise in Aleatoric \& Epistemic uncertainty}\label{gamma_noise}
We have performed another ablation study for change in uncertainty based on noise value. We estimated both aleatoric and epistemic uncertainty for visual dialog dataset. We randomly selected 200 examples on val dataset and applied noise to the image and question responses and observed that uncertainty value changes on seeing noise image and noise question. So we applied noise value $\gamma$ of 0.8 to decrease pixel value and $\gamma=1.2$ to increase pixel value i.e. there is inverse proportionality. Mean and standard deviation of uncertainties are recorded in the table. From table \ref{tab_gamma_noise_ALE_EPI}, it can be observed that aleatoric uncertainty is very small as compared to epistemic uncertainty. The aleatoric uncertainty changes much rapidly as noise changes in comparison to that of epistemic uncertainty.
\begin{figure*}[ht]
	\small
	\centering
	\begin{tabular}[b]{ c  c  c}
		(a)Distorted loss(GCE) & (b) Undistorted loss(CE) & (c) Variance Equalizer loss(VE) \\ 
		\includegraphics[width=0.33\textwidth]{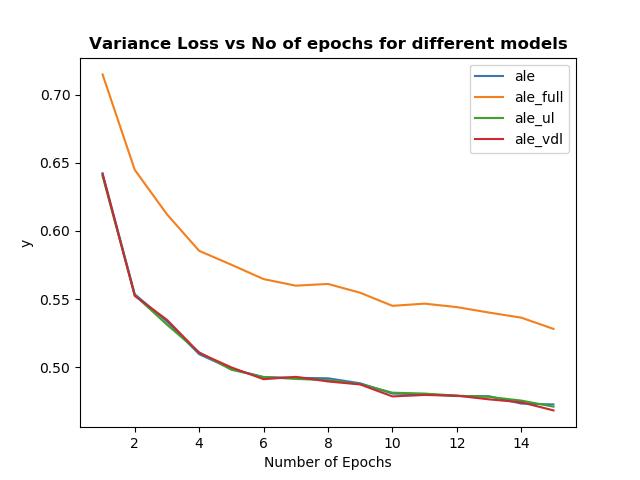}
		& \includegraphics[width=0.33\textwidth]{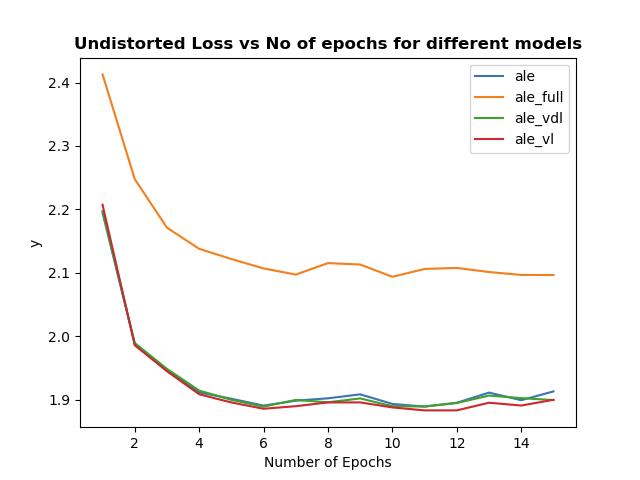}
		& \includegraphics[width=0.33\textwidth]{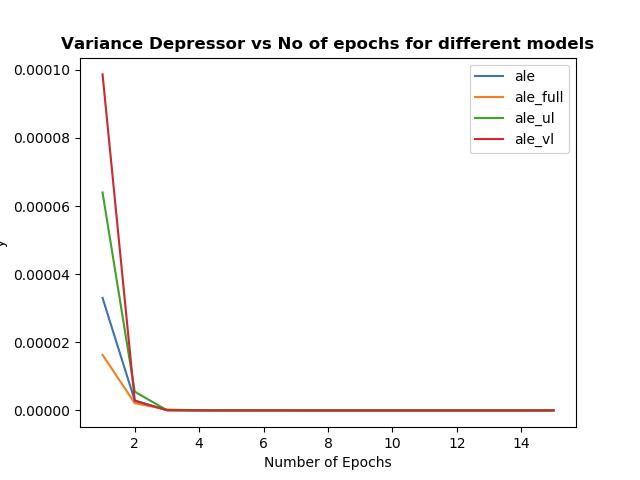}
	\end{tabular}
	\vspace{-2.2em}
	\caption{This figure shows role of different types of Losses over Epochs. From the plot we observed that variance is decreasing as it goes through more and more epochs.}
	
	\label{fig:result_sup_D}
\end{figure*}

\begin{figure*}[ht]
	\small
	\centering
	\begin{tabular}[b]{ c  c  c}
		(a)Distorted Loss(GCE) & (b) Undistorted loss(CE) & (c) Variance Equalizer loss(VE) \\ 
		\includegraphics[width=0.33\textwidth]{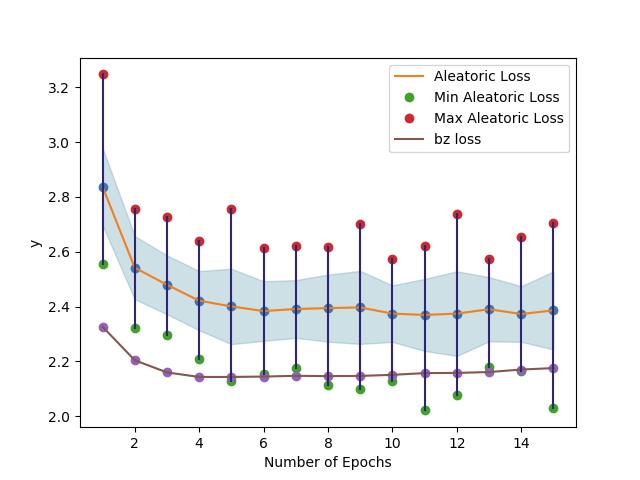}
		& \includegraphics[width=0.33\textwidth]{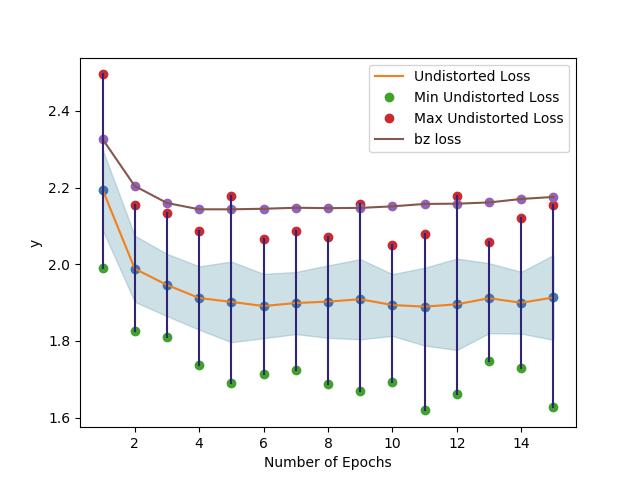}
		& \includegraphics[width=0.33\textwidth]{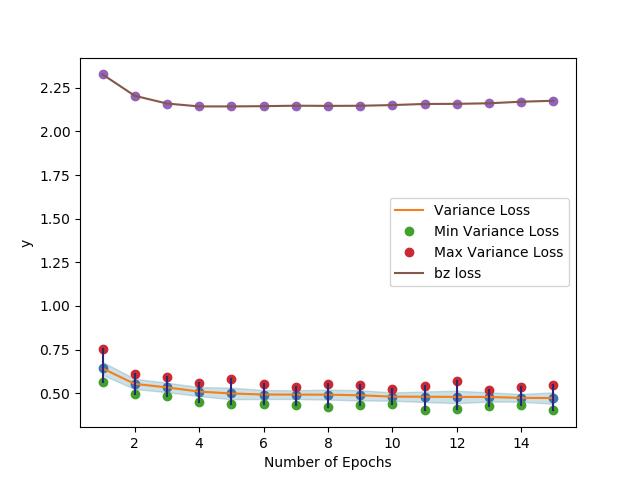}
	\end{tabular}
	\vspace{-2.2em}
	\caption{We have shown the variance  flow plots over Epochs. This shows role of different type of loss over epoch. from the plot we observed that variance is decreasing as it goes through more and more epochs.}
	\label{fig:result_sup_E}
\end{figure*}
\subsubsection{Analysis of Epistemic Uncertainty : Dropout}\label{dropout_variants}

We experimented with various dropout ratios and use the following values for the same.
For implementing Bayesian CNN, we used dropout ratio 
of (0.1, 0.2, 0.3, 0.4, 0.5) for each stack of convolutional layers respectively and 0.5 for FC layers. 
As the number of neurons increase in subsequent layers, we increase the dropout ratio for better generalization. For Bayesian LSTM, we use dropout ratio 0.3 for input \& hidden layers~\cite{Hochreiter_NC1997} and for output layer we have used 0.5 dropout ratio similar to~\cite{Gal_NIPS2016}. We further experimented with different ways of placing the dropout layer in the CNN architecture and observe that putting dropout after Max pooling layer works best.


\subsubsection{Loss Visualization}\label{Loss}
We have analyzed the significance of Distorted loss (Gaussian Cross Entropy (GCE) Loss), Undistorted loss (Cross Entropy(CE) Loss) and  Variance Equalizer (VE) loss as shown in the Figure~\ref{fig:result_sup_D}. It is clear form the figure that all the losses converges as epoch progresses.  Variance flow in the various losses is shown in Figure~\ref{fig:result_sup_E}. Also we have seen same type of behavior in the variance plot. The variance decreases for all the losses as training progresses. 


\begin{table*}[htb]
	\begin{center}
		\begin{tabular}{|l|ccccc | } \hline
			
			\textbf{Models}	 & \textbf{R1} & \textbf{R5} & \textbf{R10} & \textbf{MRR} & \textbf{Mean}\\ \hline 	
			LF \cite{Das_CVPR2017}  &43.8& 74.6& 84.0 &0.580 &5.78 \\
			HRE \cite{Das_CVPR2017}  &44.8&74.8 &84.3&0.586& 5.65\\
			MN \cite{Das_CVPR2017} &45.5& 76.2& 85.3 & 0.596&5.46\\
			HCIAE \cite{lu2017best} &48.4 &78.7 &87.5&0.622 &4.81\\
			SF-1 \cite{jain2018two}  &48.1 &78.6 &87.5 &0.620&4.79\\
			AMEM \cite{seo2017visual} & 48.5 &78.6& 87.4&0.622&4.85\\
			NMN \cite{kottur2018visual} &\textbf{50.9}& 80.1 &88.8& \textbf{0.641}& 4.45  \\\hline
			ECE  (ours)   &44.3  & 76.1  &85.9 &0.590 & 5.51\\
			ACE (ours)   &49.0  & 80.5  &89.3 &0.629& 4.32\\  
			PDUN (ours) &49.2  & \textbf{81.0} &\textbf{90.5}  &0.634 & \textbf{4.03}\\\hline
		\end{tabular}
	\end{center}
	\vspace{-1em}
	\caption{\label{score_tab_1}Results on dataset-v0.9 for Visual dialog }
\end{table*}

\begin{table}[ht]
	\centering
	\begin{tabular}{|c|c|}
		\hline
		\textbf{Model} & \textbf{Diversity($\sigma^2_d$)} \\
		\hline
		VE  & 6.41 \\
		CE  & 22.231 \\
		GCE & 27.845 \\
		ECE  &  24.980 \\
		ACE  & 32.12 \\
		PDUN & 34.35 \\
		\hline
	\end{tabular}
	\vspace{-1em}
	\caption{\label{tab_diversity} Answer Diversity results for Visual dialog.}
\end{table}
\subsection{Diversity}
We used Singular value decomposition (SVD) based evaluation metric to  demonstrate diversity across various generated answers in a dialog. Here, we have randomly selected 400 dialogs. For each dialog, we sampled $m$ number of  latent embedding feature using the attentive encoder. Each one is having $n$-dimensional feature vector. To measure the variance, all the answer embedding features can be concatenated into a feature matrix $A \in R^{m \times n}$. 
$\sigma_i$ can be obtained  by SVD, $L=U\sum V^T$ of matrix $A$, where $\sum = diag(\sigma_{0},\sigma_{1}....\sigma_{n-1})$; $U$ and $V^T$ are $m \times m$ and $n \times n$ unitary matrices respectively. The overall variance in all dimensions is $l_1$-norm, $\sigma_{o}=\sum_{i=0}^{n-1}{|\sigma_i|}$. 
A large variance indicates very less correlation among the generated answers, which further implies large diversity among the answers as shown in table~\ref{tab_diversity}.
\begin{table*}[htb]
	\begin{center}
		\begin{tabular}{|l|cccccc | } \hline
			
			\textbf{Models}	\textbf{Mean}& \textbf{R1} & \textbf{R5} & \textbf{R10} & \textbf{MRR} & \textbf{Mean} & \textbf{NDGC} \\ \hline 	
			LF \cite{Das_CVPR2017}  &\textbf{40.9}& \textbf{72.4}&\textbf{82.8} &\textbf{0.55} &\textbf{5.95}&  0.45\\
			HRE \cite{Das_CVPR2017} &39.9&70.4 &81.5&0.54& 6.41&0.45\\
			MN \cite{Das_CVPR2017} &42.4& 74.0& 84.3 & 0.56&5.59&0.47\\
			NMN \cite{kottur2018visual} & \textbf{47.5}& 78.1 &88.8& 0.61 &4.40 &0.54\\\hline
			ECE  (ours)   &43.6   & 75.4   &85.3  &0.58 & 5.36&0.49\\
			ACE (ours)   &47.0   & 82.4 &92.3  &0.62 & 3.81&0.53\\  
			PDUN (ours) &47.3   & \textbf{82.5}  &\textbf{92.6}  &\textbf{0.62} & \textbf{3.68}&\textbf{0.54}\\\hline
		\end{tabular}
	\end{center}
	\vspace{-2em}
	\caption{\label{score_tab_1_1}Results on dataset v1.0 for Visual dialog }
\end{table*}

\subsection{Comparison with state-of-the-art (SOTA)}\label{baseline_sota}
The comparison of PDUN method with various state-of-the-art methods for visual dialog dataset v0.9 and v1.0 are provided in table~\ref{score_tab_1}.  The first block of the  table~\ref{score_tab_1} consists of the state-of-the-art methods, second and third block consist of our methods. We compared our results with baseline results of the model `Late-fusion-QIH' \cite{Das_CVPR2017}. We observe that in terms of @R10 score, we obtain an improvement of around 10\% from the baseline \& 3.5\% from SOTA (NMN  ~\cite{kottur2018visual}) method. In terms of NDGC score 9\% from base model \& 0.5\% from SOTA model and in term of MRR, 7\% from the base model \& 1\% from SOTA (NMN  ~\cite{kottur2018visual}) model using our proposed method. .

\begin{figure*}[ht]
	\centering
	\includegraphics[width=1.0\textwidth]{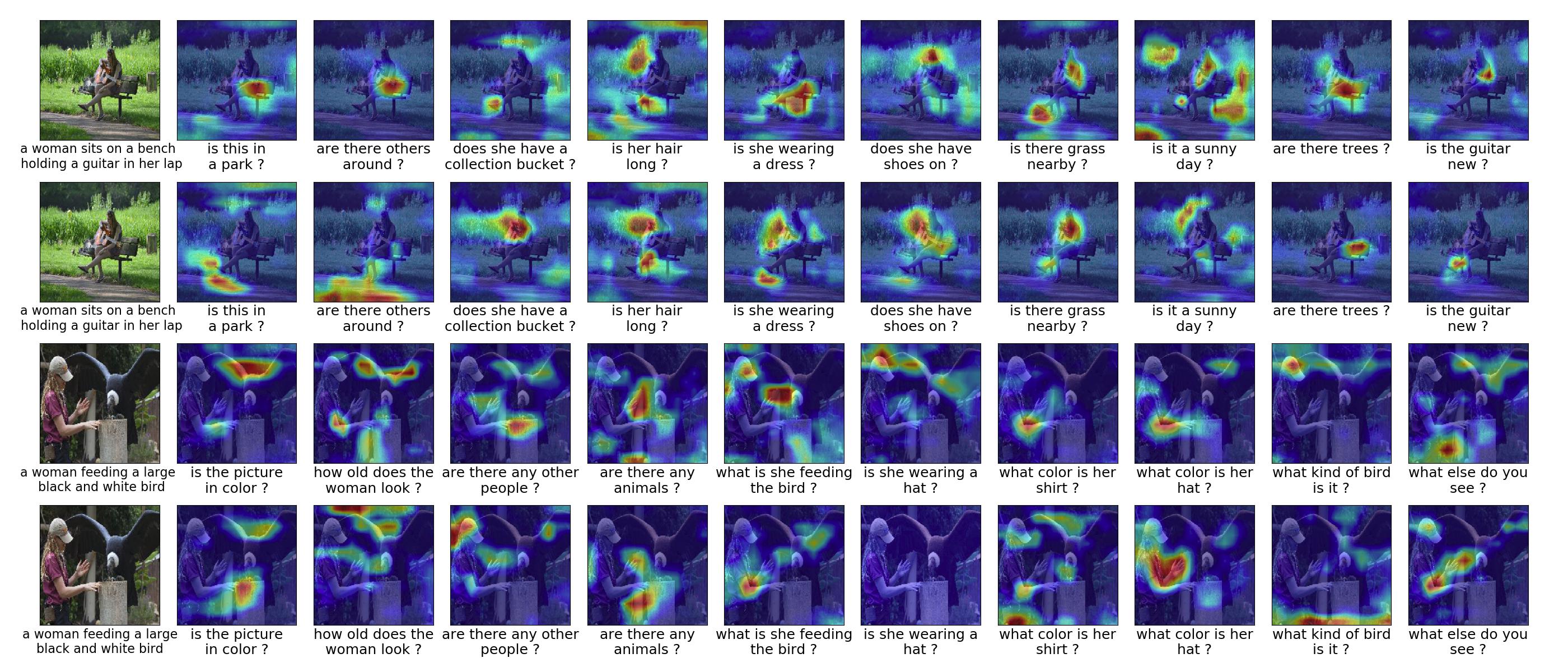}
	\vspace{-3em}
	\caption{Figure shows the difference between aleatoric dialog results and baseline dialog results. In this figure, the first row refers to Grad-CAM visualization of first example for baseline visual dialog model and second row refers to Grad-CAM visualization of first example for Aleatoric visual dialog model and same scheme is followed for next 2 rows. The first column indicates target Image and corresponding caption and starting from second column is the visualization of rounds of dialog from round 1 to 10.}
	\label{fig:result_1_C}
\end{figure*}
\vspace{-1em}
\subsection{Qualitative Result}\label{Results}
We provide qualitative results, which easily distinguishes between results of Baseline dialog model with our aleatoric dialog model for two dialog generation example in figure~\ref{fig:result_1_C}. We can clearly see that our proposed method are able to capture uncertainty and minimize it, which further improves dialog results. For example, in the first image, the question is ``Is this in a park ?". The baseline model's main focus is on the chair, where the uncertainty is very high. But our model explains about field, plant and background image, which provides the extra information about the query that eventually decreases the uncertainty as shown in figure~\ref{fig:result_1_C}. We visualize the certainty activation map  of other two dialogs whose uncertainty score decrease over turns. 

\begin{figure*}[ht]
	\centering
	\includegraphics[width=1.05\textwidth]{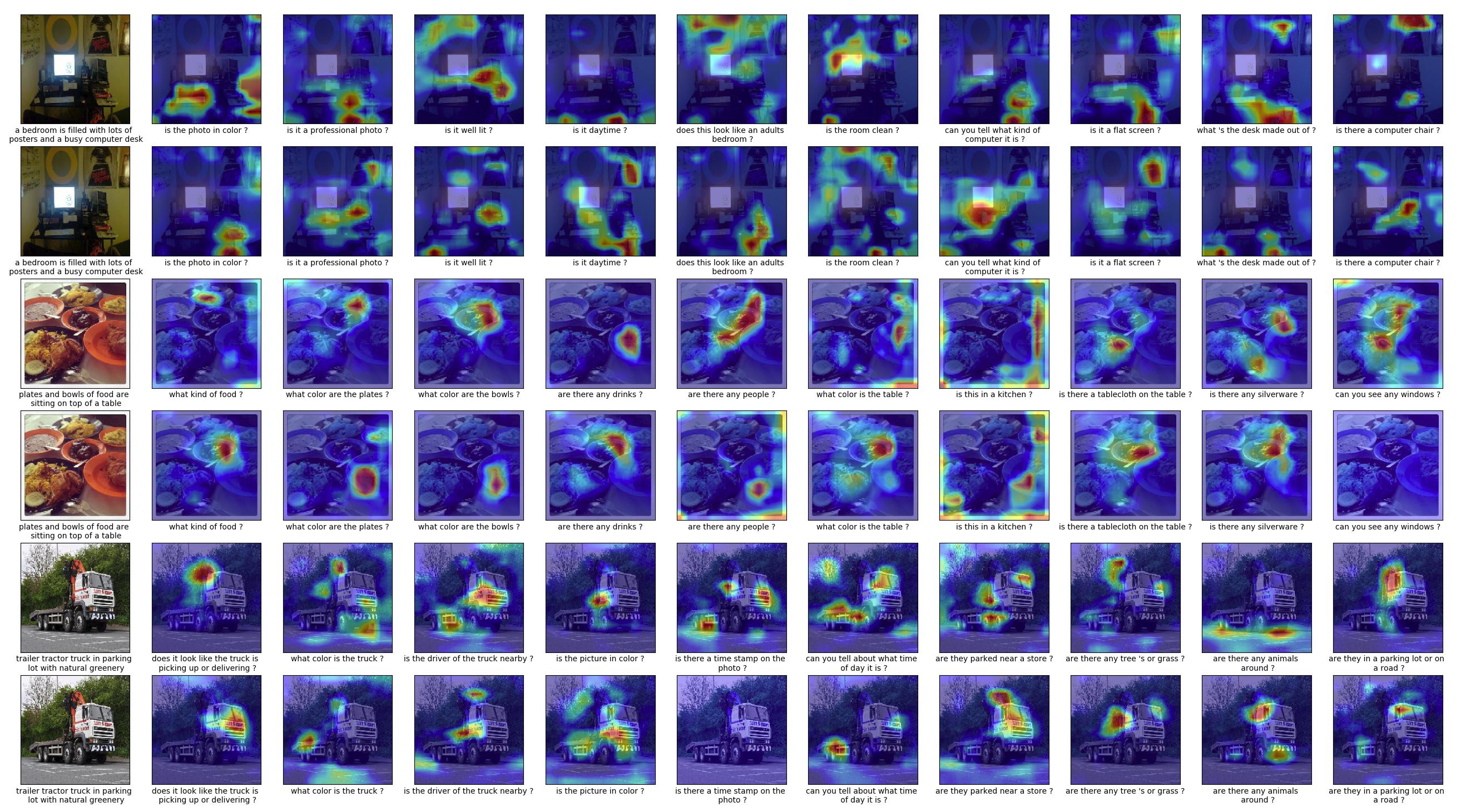}
	\vspace{-3em}
	\caption{Difference Between Aleatoric dialog results and Baseline dialog results are shown in the figure. In this figure,The first row refer to  Grad-CAM visualization of first example for  baseline visual dialog model  and second row refer to Grad-CAM visualization of first example for aleatoric visual dialog model and so on.. The first column indicates target Image and corresponding caption, second column indicates visualization of dialog round 1, third column refer to visualization of dialog round 2 and so on.}
	\label{fig:result_sup_A}
\end{figure*}

\begin{figure*}[ht]
	\centering
	\includegraphics[width=1.0\textwidth]{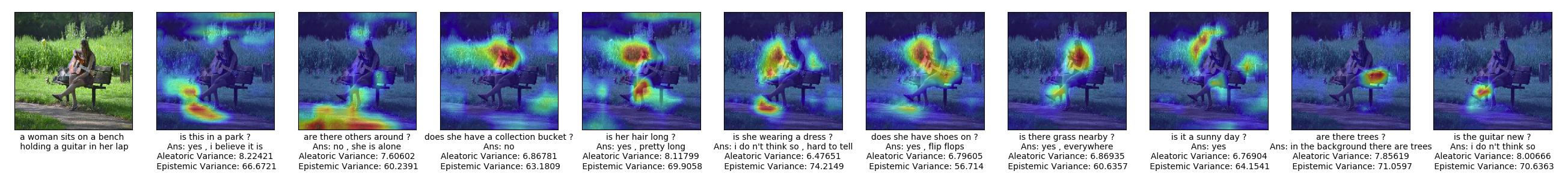}
	\vspace{-3em}
	\caption{ This figure provide aleatoric  and epidemic variance and visualize the aleatoric uncertainty using Grad-CAM for a particular Dialog. }
	\label{fig:result_sup_B}
\end{figure*}

\begin{figure*}[!htb]
	\vspace{-6em}
	\centering
	\includegraphics[width=1\textwidth]{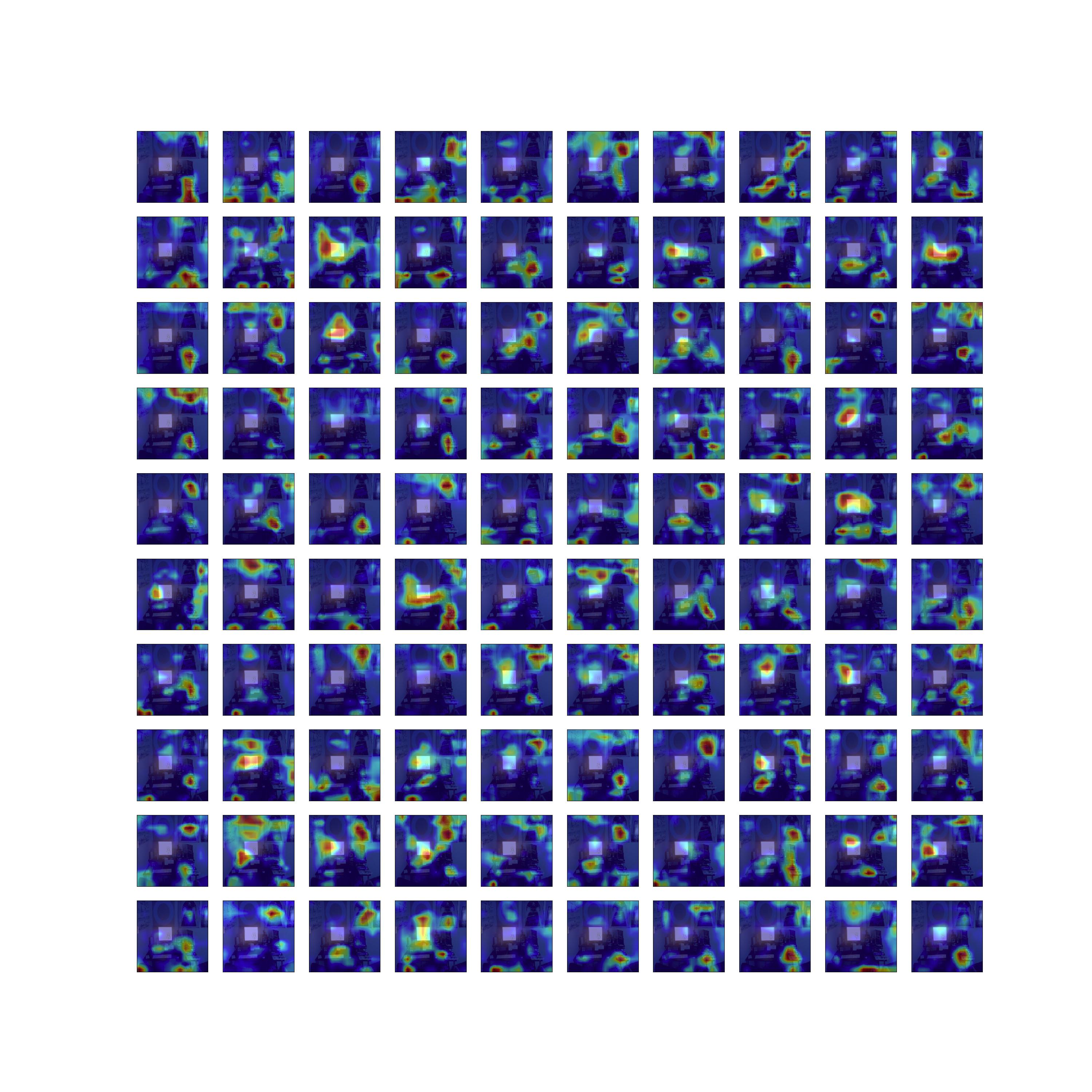} 
	\vspace{-6em}
	\caption{We visualize the multiple outputs from the Bayesian neural network. We took 100 sample from the posterior distribution of dialog model for particular image, particular question. It shows how Grad-CAM is flowing for particular image, particular question.}
	\label{fig:result_sup_C}
\end{figure*}

We provide qualitative results, which easily distinguishes between results of baseline dialog model with our aleatoric dialog model for three dialog generation example in figure~\ref{fig:result_sup_A}. We can clearly see that our proposed method is able to capture uncertainty and minimize it, which further improves dialog results. Also, we have measured epistemic and aleatoric uncertainty and showed how uncertainty decreases as dialog turns in figure~\ref{fig:result_sup_B}. We visualize the uncertainty by taking multiple samples and show how does it change as per samples as shown in Figure~\ref{fig:result_sup_C}. We also made the GIF version of this visualization with name `aleatoric\_uncertainty\_ques\_gradcam\_100.gif' and other visualisation file present in the following link\footnote{\url{https://delta-lab-iitk.github.io/PDUN/}}.

\subsection{Evaluation Protocol}
We have followed the evaluation protocol mentioned in \cite{Das_CVPR2017}. We use a retrieval setting to evaluate the responses at each round in the dialog. Specifically, every question in VisDial is coupled with a list of 100 candidate answers, which the models are asked to sort for evaluation purposes. Models are evaluated on standard retrieval metrics (1) mean rank, (2) recall @k and (3) mean reciprocal rank (MRR) of the human response in the returned sorted list.

\subsection{Preprocessing}
We truncate captions/questions/answers longer than 24/16/8 words respectively. We then build a vocabulary of words that occur at least 5 times in train, resulting in 8964 words. In our experiments, all 3 Bayesian  LSTMs are single layer with 512-dimensional hidden state. For Bayesian CNN we use pretrained VGG-19 \cite{Simonyan_arXiv2014} with dropout to get the representation of image. We first re-scale the images to $448 \times 448$ pixels and take the output of FC7 layer which is 4096-dimensional as image feature. We use the Adam optimizer with a base learning rate of 4e-4. 
\section{Conclusion}
In this paper, we propose a novel probabilistic architecture that is termed as the probabilistic diversity and uncertainty network (PDUN), for solving visual dialog. The main parts in the proposed architecture are the modules that capture uncertainty and diversity. We captured aleatoric and epistemic uncertainty that provide us with uncertainty estimates and these are further reduced using appropriate loss functions.  We have particularly shown that the performance in the visual dialog is improved around 3.5\% by the proposed network. Further the use of the diversity module obtained through a variational autoencoder allows us to generate diverse answers. We validate that indeed the diversity of the proposed network is high as compared to variants of the method. These two contributions enable us to obtain a significantly improved model for solving the challenging visual dialog task.


\section{Acknowledgment}
We acknowledge the help provided by our DelTA Lab members and our family who have supported us in our research activity.
\section{Reference}
\bibliographystyle{elsarticle-num}
\bibliography{reference.bib}

\end{document}